\documentclass[letterpaper, 10 pt, journal, twoside]{IEEEtran}

\IEEEoverridecommandlockouts                              

\usepackage{multirow}
\usepackage{siunitx}
\usepackage{makecell}
\usepackage{amsmath} 
\usepackage{amssymb}  
\usepackage{booktabs}
\usepackage{color}
\usepackage[pdftex]{graphicx}
\usepackage{subfigure}
\usepackage{microtype}
\usepackage{fancyhdr}

\input{abkuerzung.sty}
\graphicspath{{../pdf/}{../jpeg/}}
\DeclareGraphicsExtensions{.pdf,.jpeg,.png}

\renewcommand{\baselinestretch}{1}

\begin{document}

\markboth{IEEE Robotics and Automation Letters. Preprint Version. Accepted January, 2022} {Kirschner \MakeLowercase{\textit{et al.}}: Expectable Motion Unit}  

\author{Robin~Jeanne~Kirschner, 
        Henning~Mayer,
        Lisa~Burr,
        Nico~Mansfeld,
        Saeed~Abdolshah
        and Sami~Haddadin 
\thanks{Manuscript received: September, 9, 2021; Revised December, 3, 2021; Accepted January, 2, 2022.}
\thanks{This letter was recommended for publication by Angelika Peer upon evaluation of the Associate Editor and Reviewers' comments.}
\thanks{This work was supported by the European Union's Horizon 2020 research and innovation programme as part of the projects Darko (grant no. 101017274) and I.AM. (grant no. 871899). We gratefully acknowledge the funding of the Lighthouse Initiative Geriatronics by StMWi Bayern (Project X, grant no. 5140951) and LongLeif GaPa gGmbH (Project Y, grant no. 5140953) and the support by the Bavarian Institute for Digital Transformation.}
\thanks{Robin~Jeanne~Kirschner, Lisa~Burr, Nico~Mansfeld, Saeed~Abdolshah, and Sami~Haddadin are with the Munich Institute of Robotics and Machine Intelligence (MIRMI), formerly MSRM, Technical University of Munich, Munich, Germany. Henning Mayer is with the School of Social Sciences and Technology, Technical University of Munich, Munich, Germany, {\tt\footnotesize robin-jeanne.kirschner@tum.de}.}%
\thanks{Digital Object Identifier (DOI): 10.1109/LRA.2022.3144535.}
}

\title{Expectable Motion Unit:
Avoiding Hazards From Human Involuntary Motions in Human-Robot Interaction
}

\maketitle

\pagestyle{fancy}
\fancyhead{} 
\fancyfoot{} 
\fancyfoot[LE,RO]{\thepage}           
\fancyfoot[RE,LO]{\fontsize{6}{8} \selectfont © 2022 IEEE.  Personal use of this material is permitted.  Permission from IEEE must be obtained for all other uses, in any current or future media, including reprinting/republishing this material for advertising or promotional purposes, creating new collective works, for resale or redistribution to servers or lists, or reuse of any copyrighted component of this work in other works. DOI: 10.1109/LRA.2022.3144535}


\begin{abstract}
In robotics, many control and planning schemes have been developed to ensure human physical safety in human-robot interaction. The human psychological state and the expectation towards the robot, however, are typically neglected. Even if the robot behaviour is regarded as biomechanically safe, humans may still react with a rapid involuntary motion (IM) caused by a startle or surprise. Such sudden, uncontrolled motions can jeopardize safety and should be prevented by any means. In this letter, we propose the Expectable Motion Unit (EMU), which ensures that a certain probability of IM occurrence is not exceeded in a typical HRI setting. Based on a model of IM occurrence generated through an experiment with 29 participants, we establish the mapping between robot velocity, robot-human distance, and the relative frequency of IM occurrence. This mapping is processed towards a real-time capable robot motion generator that limits the robot velocity during task execution if necessary.
The EMU is combined in a holistic safety framework that integrates both the physical and psychological safety knowledge.
A validation experiment showed that the EMU successfully avoids human IM in five out of six cases.
\end{abstract}


\section{INTRODUCTION}
\IEEEPARstart{S}{afety} is a key requirement for successfully implementing modern collaborative and tactile robots in real-world industrial and service scenarios. As proximity is an essential part of human-robot interaction (HRI), collisions and contact (desired, undesired, or even unforeseen) may occur. In robotics, many pre- and post-collision strategies have been introduced to ensure the human physical integrity, e.g., collision detection and reaction \cite{Haddadin_2017}, collision avoidance \cite{Pereira_2019}, \cite{Weitschat_2018}, and real-time model-, metrics-, or injury data-based control \cite{Rossi_2015, Tadele_2014}.
Besides ensuring the human's physical integrity, safe and efficient human-robot interaction also requires consideration of the human physical state, psychological state, and responses, e.g., the human pose \cite{mainprice_2013}, \cite{liu_2017}, \cite{Lasota_2015}, the affection towards robots \cite{rani_2004}, or the anxiety caused by a robot \cite{Kulic_2005}. An important factor to be considered in HRI is the human expectation \cite{Knepper_2017}. If it is violated, then humans may react with a startle or surprise reaction \cite{Lange_2018}. This includes rapid involuntary human motions (IM), which may jeopardize safety \cite{Martin_2015}.

To avoid possibly hazardous contacts, several authors considered the worst case human motion range and dynamics in the development of safe control and planning schemes \cite{Weitschat_2018, Pereira_2019}. Such schemes can become overly conservative and may lead to large separation distances between the human and the robot, even if the probability of a rapid motion such as IM is low.
\begin{figure} 
	\centering
	\includegraphics[width=\columnwidth]{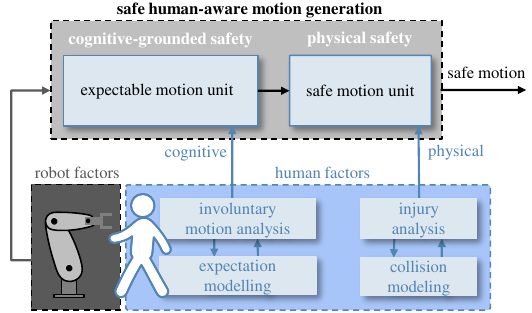} 
	\caption{Proposed framework for safe human-aware motion generation that combines cognitive-grounded safety aspects and well-established physical safety considerations and paradigms. The main contribution of this work is an experimental model of human involuntary motion (IM) occurrence in HRI and the derivation of the Expectable Motion Unit (EMU), which avoids potentially dangerous human IM in HRI.}
	\vspace{-5mm}
	\label{fig:intro}
\end{figure}
Safer, closer, and more efficient HRI can be realized if human IM occurrence (IMO) is reduced or even excluded.

In this letter, we introduce a systematic approach to improve performance and safety in HRI by avoiding human IM. First, we investigate the influence of robot motion parameters on the probability of human IMO in a common use case via an exploratory study involving 29 participants. The collected data and knowledge are then processed towards a human-aware, real-time capable motion generator that limits the robot speed so that a certain IMO probability is not exceeded. This tool is called \emph{Expectable Motion Unit (EMU)}; see Fig. \ref{fig:intro}. It can be seamlessly combined with state-of-the-art safety schemes for avoiding human injury or pain during collisions, such as the Safe Motion Unit (SMU), which was proposed in \cite{Haddadin_2012}. A preliminary validation experiment involving eleven participants shows that the proposed framework successfully avoids IM in the considered use case. Overall, by taking both the human physical safety and expectation into account, our concept ensures safety in practice and improves the acceptance and trustworthiness in HRI.  

\begin{figure*} [htp!]
	\centering
	\includegraphics[width=\textwidth]{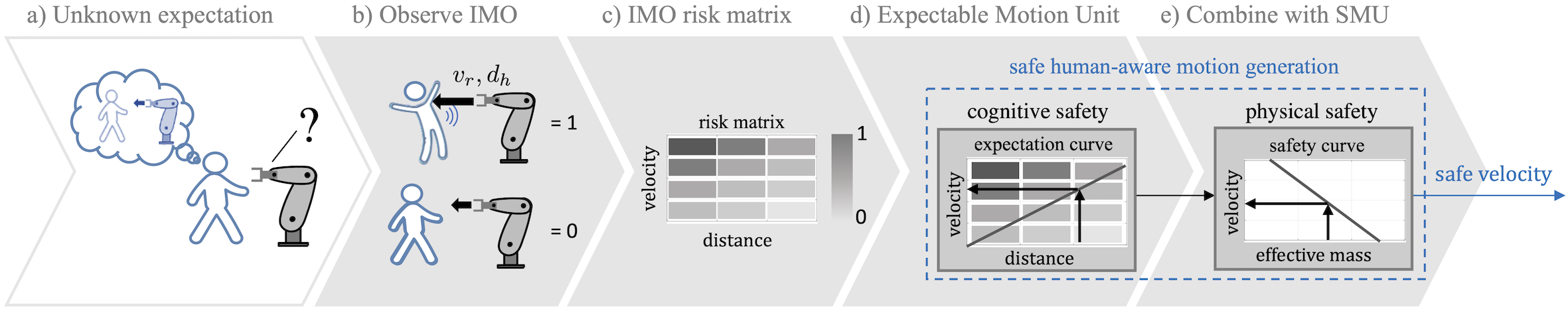}
	\caption{Derivation of the EMU. The first step is to experimentally investigate under which circumstances (velocity of the robot $v_\mathrm{r}$ and distance to the human $d_\mathrm{h}$) human IMO occur (b). Using the probability of IMO (between 0 and 1) resulting from the experiment, an IMO risk matrix is assembled (c). The risk matrix allows us to define expectation curves, that can be embedded into the EMU (d), which is finally combined with the SMU to ensure both physical and perceived safety.}
	\label{fig:pipeline_approach}
\end{figure*}

The remainder of this letter is organized as follows. In Sec. \ref{sec:sota}, we provide a brief overview of related work. In Sec. \ref{sec:approach}, we introduce the EMU. The exploratory experiment and evaluation method to derive the risk of IMO are described in Sec. \ref{sec:meth}. The implementation of the EMU is treated in Sec. \ref{sec:EMUimp}, followed by a validation experiment. Sec. \ref{sec:disc} discusses the results and future directions of research. Finally, Sec. \ref{sec:conc} concludes the letter.

\section{RELATED WORK}
\label{sec:sota}
This section briefly summarises related work on physical safety and the human perception of safety in close HRI, particularly the influence of the robot velocity and human-robot distance.

A central goal in HRI is to ensure that no injury occurs during wanted or unwanted contact \cite{ISO_TS}. Several studies were conducted to understand the injury mechanisms of different human body parts in direct collisions \cite{Haddadin_2012,haddadin_icra2009_part1, haddadin_icra2009_part2, haddadin_et_al_ram2008,behrens2019biomechanische}. For ensuring safety via control, many model- and metrics-based schemes have been proposed \cite{Rossi_2015, Tadele_2014,ISO_TS,Lacevic_2010}. However, a drawback of such approaches is that contact models can typically not capture the complex human injury mechanisms. The authors of \cite{Haddadin_2012} proposed a data-driven, model-independent approach that relates medically observed injury directly to collision ``input'' parameters instead of using physical ``output'' quantities such as force or pressure. More specifically, the robot effective mass in the direction of motion \cite{Khatib_1995}, the robot velocity, and the surface curvature are mapped to injury severity. This mapping is embedded into the so-called Safe Motion Unit (SMU), which limits the robot velocity along the desired trajectory to a biomechanically safe value.


Besides ensuring physical safety, several authors embedded psychological aspects such as human comfort into robot motion control \cite{Sarkar_2002, Alami_2011, Cao_2018}. Comfort is associated (among other human factors) with the user's expectation of the robot behavior. If the expectation is not fulfilled, then startle and surprise can cause users to react with IM, which may be potentiated by fear \cite{Martin_2015}.
Literature on how startle and surprise can be identified and how they relate to each other is diverse and requires context-dependent analysis. According to \cite{Tomkins_1962}, both startle and surprise lie on a continuum where startle is a strong surprise reaction. Both types of responses can negatively impact safety in HRI applications and can be observed through multimodal social signal coding. Social cues that follow unexpected robot movements are, e.g., felt smiles \cite{Ekman_1982} or body freezes \cite{Coulson_2004}. The human expectation shapes the reaction towards a robot motion in a certain event and is influenced by the human perception of the scenario \cite{Lange_2018}.

Research on the psychological state of the human in HRI mainly considers the perception of human safety \cite{Bartneck_2009, Kulic_2005} and influencing factors. The human-robot distance was shown to influence the perception of safety and comfort in HRI \cite{Walters_2005, Takayama_2009, Mumm_2011}, which proxemic behaviour models can explain.
The proxemics model was introduced in \cite{Hall_1966} and describes the social zones in which humans feel comfortable interacting with each other. The authors of \cite{Walters_2005, Takayama_2009, Mumm_2011}, investigated which  robot-human distance is preferred by the human. 
Unexpectedly, in studies with the mechanical looking \emph{PeopleBot}, it was found that $40~\%$ of the participants were comfortable with the robot in the proximity of less than \SI{0.45}{m}, where humans usually only accept intimate relationships. The fact that even interaction distances of less than \SI{0.45}{m} are still acceptable for users suggests that not everyone considers the robot as a social instance \cite{Walters_2005}. Table~\ref{social_zones} lists the human-robot distances that were accepted by the participants in \cite{Walters_2005}, \cite{Takayama_2009}, \cite{Mumm_2011} compared to the well-established proxemics model \cite{Hall_1966}. In addition to the distance between the human and the robot, several studies conclude that the robot velocity is a factor that strongly correlates with the feeling of arousal \cite{Kulic_2005}, perceived level of hazard \cite{Or_2009}, and perceived safety \cite{Nonaka_2004}. The authors of \cite{Dehais_2011} also investigated the influence of robot approaching motion parameters, including varying motion direction, acceleration, and jerk in the context of human-robot handover scenarios.



\begin{table} [t]
	\centering
	\caption{Social Zones for Robot Acceptance regarding ``Proxemics''}
	\label{social_zones}
	\begin{tabular}{c c c}		
		\toprule
		Personal space zone & Range & Reference\\	
		\midrule
		Close intimate & $0.00$ m - $0.15$ m & \cite{Walters_2005} \\
		Intimate  & $0.15$ m - $0.45$ m & \cite{Walters_2005}, \cite{Takayama_2009} \\
		Personal & $0.45$ m - $1.20$ m & \cite{Walters_2005}, \cite{Takayama_2009}, \cite{Mumm_2011} \\
		Social & $1.20$ m - $3.60$ m \\
		Public & $\ge 3.60$ m &  \\
		\bottomrule
	\end{tabular}
	\vspace{-2mm}
\end{table}

\section{EXPECTABLE MOTION UNIT CONCEPT}
\label{sec:approach}
In this work, we propose the Expectable Motion Unit (EMU), a cognitive-grounded safety concept based on human expectation fulfillment. The EMU aims to ensure that human expectation is fulfilled at any time. This avoids startle and surprise reactions, which can cause possibly hazardous human IM. Based on a model of IMO obtained through a participant study, the EMU reduces the robot velocity during task execution when necessary; see Fig. \ref{fig:intro}. The EMU closes the gap between online-planning methods for psychological aspects for human safety in HRI and physical safety considerations and paradigms \cite{Lasota_2017}. The derivation of the concept is described in the following; see Fig. \ref{fig:pipeline_approach}.




The first step is to understand under which circumstances IM occur in HRI. Obviously, many human, environmental, and robot factors influence human expectation and thus human IMO. In \cite{Hall_1966, Kulic_2005, Takayama_2009, Bartneck_2009}, it was shown that the robot velocity and inter-subject distance are two major factors that affect human perceived safety. Assuming perception correlates with expectation as previously shown in \cite{Lange_2018}, we are interested in how these two parameters also influence IMO. Therefore, in this work, we systematically investigate the influence of the robot speed and the robot-human distance on IMO for a common class of industrial HRI tasks. In this task,
\begin{itemize}
    \item the robot approaches the human upper body frontally,
    \item the human focuses on her/his task,
    \item the human has a certain mental occupancy, and
    \item the human is aware that the robot is moving and sharing the same workspace.
\end{itemize}
Many common assembly or transportation tasks belong to this class. We assume that the human has a certain task-dependent expectation of the robot's behavior when working in close proximity. For example, the human may expect that the robot moves slowly inside the human's workspace.
%
%
To observe IMO under the mentioned conditions, the robot approaching motion is performed with variable speed and distance; see Fig. \ref{fig:pipeline_approach} a)-b). The human reaction is recorded and classified via social cue analysis. From the experiments, we derive the relative frequency of IMO depending on the robot velocity and robot-human distance. We call this mapping a \emph{risk matrix} for IMO; see \ref{fig:pipeline_approach} c). We can then define a threshold of IMO probability that shall not be exceeded. In the risk matrix, this threshold can be represented by a so-called \emph{expectation curve}, which relates the current human-robot distance to a robot velocity that meets the human expectation; see Fig. \ref{fig:pipeline_approach} d). The expectation curve is integrated into the EMU, which limits the robot speed (if necessary) to a value that is considered to be expectable for the human. Finally, the EMU is combined with the Safe Motion Unit (SMU), which provides a safe velocity based on injury data from biomechanics collision experiments; see Fig. \ref{fig:pipeline_approach} e). The output of the framework is a velocity that ensures physical safety and reduces IM simultaneously.



\section{EXPERIMENTAL OBSERVATION OF IMO}
\label{sec:meth}
This section covers the experimental observation of human IMO in a common HRI scenario. We first describe the experimental design and procedure. Then, the social signal coding for the IMO evaluation is explained. Finally, we provide results on the coder reliability for the social signal analysis. Our experiments were conducted under the approval of the Ethics Commission of Technical University of Munich (TUM) under review no. 395/19 S.

\begin{table} 
	\caption{Summary experimental conditions for the observation of IMO.}
	\label{tab:exp}
	\centering
	\begin{tabular}{l l l }
		\toprule
		
		
		 \makecell[l]{\textbf{Objective}} &  \multicolumn{2}{c}{\makecell{Generate data for\\ expectation curves using \\ maximum reachable \\ velocities}} 
		 \\
		\midrule
		\makecell[l]{\textbf{Approach}\\ \textbf{distance} to\\ tablet's edge \\ (see Fig. \ref{fig:exp_proc} a):} &  \multicolumn{2}{c}{\makecell{$ d_\mathrm{h1}=0.00$ m to  $ d_\mathrm{h6}=0.25$ m \\ Steps: $ \Delta d_\mathrm{h}= 0.05$ m\\ Order: randomized}} \\
		\midrule			
		\makecell[l]{\textbf{Approach}\\ \textbf{velocity}} &  \makecell[l]{Set 1: \\$0.25$ m/s} &  \makecell[l]{Set 2: \\$0.55$ - $1$ m/s\\ Steps: $0.05$ m/s\\ Order: lower \\ $v_r$ with\\higher $d_h$} 
		\\
		\midrule
		\makecell[l]{\textbf{Observed}\\ \textbf{parameter}} &  \multicolumn{2}{c}{\makecell{S-S cues}} \\
		\midrule								  
		  \makecell[l]{\textbf{Participants}} & \multicolumn{2}{c}{\makecell{number: 29 \\ age: $34.3$ ($\pm15.9$), \\male: 21 ($72.4\%$), \\female: 8 ($27.6\%$)}} 
		  \\ 
		\bottomrule

	\end{tabular}
	\vspace{-5mm}
\end{table}

\begin{figure*} 
	\centering
	\includegraphics[width=2\columnwidth]{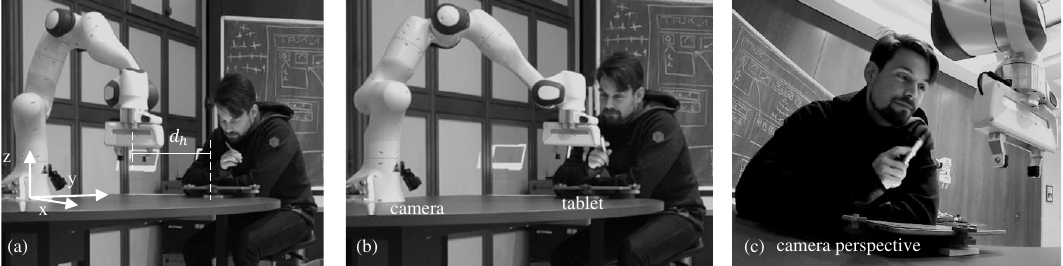}
	\caption{Experimental setup and procedure: (a) The participant focuses on a quiz while the robot moves at some distance. (b) The robot approaches towards distance $d_h$. (c) The camera captures mimics and gestures.}
	\label{fig:exp_proc}
	\vspace{-5mm}
\end{figure*}

\subsection{Experimental Procedure and Design}
\label{sec:exp_proc}
The experimental setup introduced in \cite{Kirschner_2021_ps} is used; see Fig. \ref{fig:exp_proc}. It consists of a Franka Emika Panda robot manipulator that is mounted on a table, a PC, a camera that captures the human upper body and face, a tablet placed 
at $\SI{0.44}{m}$ distance from the robot base in the y-direction, and a standing-chair that ensures all participants' heads are positioned at the same height; see Fig. \ref{fig:exp_proc} a). 

We conduct two sets of experiments, which are summarized in Tab. \ref{tab:exp}. In both sets, equally spaced robot-human distances in the range \SI{0}{m} to \SI{0.25}{m} were selected. In the first set, the robot velocity is set to \SI{0.25}{m/s}, which is the current velocity limit for safe teach-in according to ISO/TS 15066:2016. In the second set, both the velocity (\SI{0.55}{}--\SI{1}{m/s}) and the distance are varied.
 

The human-robot distance is defined such that the perceived safety according to \cite{Takayama_2009} and \cite{Walters_2005} is low, which means that IM are likely to occur. We start with the maximum distance $d_\mathrm{h6} = \SI{0.25}{m}$, which was perceived as comfortable for more than \SI{25}{\%} of participants in \cite{Walters_2005}. The distance was then successively decreased to $d_\mathrm{h1}=\SI{0}{m}$ in \SI{5}{cm} steps. 
The other experimental conditions are the same for both sets. The experiment consists of three parts. In the first part, the participant is seated in front of the tablet. He/she is asked to complete a quiz about the safe use of the robot arm on the tablet, which requires a certain mental occupancy. The quiz task consists of
\begin{itemize}
\item[1.]{ a short fixation time to analyze the problem (measured fixation time among three volunteers: 8.21$\pm$4.38 s) followed by} 
\item[2.]{ a consideration time (e.g., for observing the environment) and}
\item[3.]{ a hand-eye coordination task, i.e., tapping on the correct answer.} 
\end{itemize}

The participants are informed that the robot is moving while they are solving the quiz. The robot motion is then initiated, and the manipulator moves within a randomized number of squares in the x/y-plane at a distance of $d_\mathrm{h} \geq $ \SI{0.44}{m} to the tablet. Then, the robot end-effector approaches the human workspace after a randomized time and stops in front of the tablet at one of six distances ranging from $d_\mathrm{h} = $ \SI{0.25} to \SI{0.00}{m} in a randomized order; see Fig. \ref{fig:exp_proc} b). The participant's reaction is recorded by a camera mounted at the robot base; Fig. \ref{fig:exp_proc} c) shows an exemplary camera image. 

The participants are asked to answer a Godspeed V questionnaire to rate their perceived safety \cite{Bartneck_2009} directly after the robot approaching motion.  
The participants are aware that the robot moves at some distance, but not that it approaches them at a certain point in time. Thus, the participants do not expect the first approach. After the first approach, the participants are aware that the robot may approach them again.

As most participants were inexperienced with the robot system (or any other robot), part two of the experiment serves as training and habituation and consists of hands-on training in groups of three persons. It starts with an introduction to the robot and its programming interface, followed by hands-on programming of a pick and place task. After a short break, the group is split up again, and for the third experimental part the participants are individually asked to sit down in front of the tablet to complete another quiz. The previous experiment (part one) is repeated for every participant.

All participants were asked to fill in a demographic questionnaire and a Technology Usage Inventory in addition to the Godspeed questionnaire to investigate possible human factors that influence IMO. A thorough investigation of the influence of different human factors on IMO is not in the scope of the this letter but will be addressed in future work.

\subsection{Social-Signal Analysis for Involuntary Motion}
\label{subsec:ssprocessing}
In the experiment, we aimed at creating a natural environment for the human participants. Therefore, we excluded intrusive measurement methods for detecting human arousals, such as EEG sensors attached to the human face or forehead. For an objective measurement of startle or surprise, we asked the participants to wear a Garmin Forerunner 645 Music smartwatch with heart rate (HR) and heart rate variability (HRV) measurement function as we considered wearing a watch as natural to most people. 

Based on the assumption that IM can be indicated by social cues of startle and surprise (S-S) as suggested in \cite{Kirschner_2021_ps}, we conduct a multi-modal video analysis (facial displays, gaze, gestures/postures), in which S-S reactions are recorded and labelled. Even though computational approaches for the automatic recognition of social signals have been proposed, manual annotation procedures are currently better suited to identify and interpret situated expressions in our collaboration task \cite{Pitsch_2016}. In this article, we use a context-sensitive approach, where the expert evaluates the social signals (human coder) as an interpretation filter. Based on theoretical assumptions and sample training, he/she decides whether a given cue can be classified as a social S-S signal. Both inductive sample training and deductive theoretical assumptions based on pre-existing knowledge about S-S cues form the basis of our codebook. To achieve a high degree of validity and generalisability of the video analysis, we let two human coders evaluate the video files independently based on the deductive and inductive codebook.
The first and main coder is male, has a sociological background, and has experience in social cue coding. To ensure that his bias is as small as possible, the coder is given the experimental videos for evaluation, knowing only the general setup but not the purpose of the study. He is asked to prepare a codebook and to evaluate the videos concerning the frequency and strength of S-S reactions. The second coder is a female mechanical engineer who has no previous social cue coding experience. She knows the goal of the experiment and is involved in the implementation and execution of the experiment. 
To ensure high reliability in terms of individuality and temporal consistency of the annotation, inter- and intra-coder reliability checks are carried out: 
\begin{itemize}
	\item The inter-coder reliability is ensured by using two coders with different levels of experience. Both coders analyse all videos of the experiments and apply the same coding instrument. Then, their level of agreement is determined.
	\item The intra-coder reliability is ensured by repetition of the analysis two months later by the main coder and generating the Cohens Kappa score among the annotations \cite{Landis_1977}.
\end{itemize}

The codebook is guided by the scheme listed in Tab. \ref{tab:scheme}. It was derived from well-known literature \cite{Coulson_2004, Grillon_2003, Davis_1984}. During sampling training (two videos with a duration of 13 min 53 s in total), we identified an additional expression, namely felt smiles \cite{Ekman_1982}. We assume them to be relief reactions that follow the S-S responses. Felt smiles were observed in participants after unexpected robot movements. For the annotation of non-verbal behaviour, we use a simplified version of the MUMIN multimodal coding scheme proposed in \cite{Allwood_2007} and adapt it by including gaze and body postures from human users and removing parts concerning human-human interaction. The ELAN annotation tool \cite{Wittenburg_2006} is used to classify the videos. The coder watches the entire video clip of one participant running through all experimental conditions as a first step. Then, the baseline postures, gestures, and facial displays, including gaze, are annotated at the beginning of the video sequence, when the robot did not perform the approach yet. Subsequently, only those changes in the expressive behaviour of human participants are annotated, that
\begin{itemize}
	\item{can be associated with movements of the robot arm by registering basic feedback of contact perception and}
	\item{indicate that the human is startled and/or surprised by these movements.} 
\end{itemize}

\begin{table} 
	\caption{Coding scheme and respective references for social cues on startle and surprise}
	\label{tab:scheme}
	\centering
	\begin{tabular}{l l l l l l l |}
		\toprule
		& \textbf{Facial display} & \textbf{Gestures/Postures} \\	
		\midrule
		
		\makecell[l]{\textbf{Startle}\\ (reflex)} &  \makecell[l]{- rapid eyeblinks [RE] \cite{Grillon_2003}\\- lowered eyebrows [LE] \cite{Ekman_1985}\\- closed eyes [CE] \cite{Ekman_1985}\\- tightened eyelids [TE] \cite{Ekman_1985}\\- horizontally stretched\\    lips [HSL] \cite{Ekman_1985}\\- tightened neck [TN] \cite{Ekman_1985} \\- delayed felt smile \\(relief) [DFS]} &  \makecell[l]{- evasive head \\ movements [EHM] \cite{Davis_1984}\\- evasive trunk\\ movements [ETM] \cite{Coulson_2004}\\- shoulder jerks [SJ]\\- body twitches [BT]\\- body freezes [BF] \cite{Coulson_2004} }\\
		\midrule											  
		\makecell[l]{\textbf{Surprise} \\ (emotion)} & \makecell[l]{- raised eyebrows [REB] \cite{Ekman_1985}\\- widened eyes [WE] \cite{Ekman_1985}\\- raised upper \\eyelids [RUE] \cite{Ekman_1985}\\- open jaws [OJ] plus relaxed\\ lips [RL] \cite{Ekman_1985}} & \makecell[l]{\\- evasive head\\ movements [EHM] \cite{Davis_1984}\\- evasive trunk\\ movements [ETM] \cite{Coulson_2004}\\- body freezes [BF] \cite{Coulson_2004} }\\
		\bottomrule

	\end{tabular}
	\vspace{-2mm}
\end{table}

Possible cues for contact perception include immediate or delayed gaze changes towards the robot and interruptions of the human task, which are indicated, e.g., by freezing hand gestures. The coder reports that no IM occurred if no S-S cue occurred.

\subsection{Reliability of the Social Signal Coding}
In order to check the reliability of the social signal coding regarding individuality and temporal consistency, we calculate the inter- and intra-coder reliability as explained previously. The reliability of the evaluation in terms of the Cohens-Kappa score ($\kappa$) \cite{Landis_1977} is
\begin{itemize}
    \item {inter-coder reliability: $\kappa =$ 0.805}
    \item {intra-coder reliability: $\kappa =$ 0.840}
\end{itemize}
This can be considered as ``almost perfect'' according to \cite{Landis_1977}.

We did not observe a significant change in HR or HRV when the participants showed S-S cues. Therefore, we excluded the measurement from further evaluation.
\section{SAFE HUMAN-AWARE MOTION GENERATION}
\label{sec:EMUimp}
In this section, we describe how the EMU can be implemented using the data collected in the previous experiment. First, we derive the so-called risk matrix and expectation curves. The expectation curves are then embedded into the EMU, which is implemented on the Franka Emika Panda in combination with the SMU \cite{Haddadin_2012}. Finally, the EMU is validated in a further participant study. 

\subsection{Risk Matrix Identification}
\label{sec:results}
\begin{figure}[t] 
	\centering
	\includegraphics[width=\columnwidth]{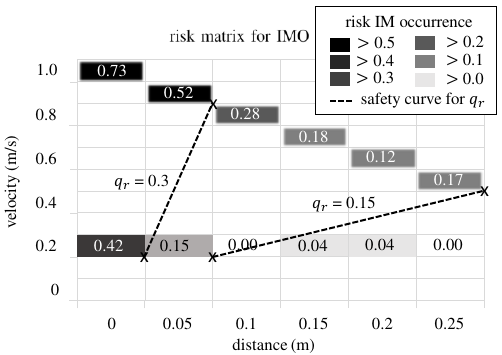}
	\caption{Exploratory velocity-distance risk matrix for IMO with two exemplary expectation curves for the IM thresholds $q_r = 0.15$ and $q_r = 0.30$. Depending on the human-robot distance, the expectation curves provide a robot velocity that avoids IM with the selected probability $q_r$.}
	\label{fig:s_curve} 
\end{figure}

\begin{figure} 
	\centering
	\includegraphics[width=\columnwidth]{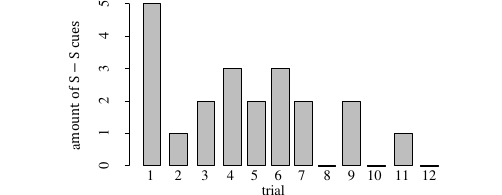}
	\caption{Total number of S-S cues for all participants over the number of robot approaching motions.}
	\label{fig:dist_cues}
	\vspace{-5mm}
\end{figure}

We summarize our experimental results in a risk matrix, which maps the human-robot distance and velocity to a probability of IMO; see Fig.~\ref{fig:s_curve}.
In our experiments, we observed a significantly higher number of S-S cues in the first approach compared to the following ones; see Fig. \ref{fig:dist_cues}. From this, we conclude that the participants do not expect the first robot approach motion but are aware of the subsequent approaches, i.e., there is a change in situational awareness. Therefore, we exclude the results of the first approach from the calculation of the risk matrix.


In the risk matrix illustrated in Fig. \ref{fig:dist_cues}, the IMO probability is listed for the two experimental sets, i.e., a) constant velocity (\SI{0.25}{m/s}) and variable distance (horizontal entries) and b), variable distance and variable robot velocity (diagonal entries).
Notably, the IMO probability decreases below $d_\mathrm{h} =$ \SI{0.10}{m}, which most likely results from the low number of participants. From the risk matrix, we can deduce an \emph{expectation curve} for a certain threshold $q_r$ in terms of IMO probability. With this expectation curve, we can determine the maximum robot velocity that can be commanded while satisfying the IMO constraint.
Two exemplary expectation curves\footnote{In this letter, the expectation curves are linear for the sake of simplicity. However, in general, they may have another, e.g., exponential, shape.} for $q_r = $ \num{0.3} (\SI{30}{\%}) and \num{0.15} (\SI{15}{\%}) are illustrated in Fig.~\ref{fig:s_curve}.

\begin{figure}[t]
	\centering
	\includegraphics[width=\columnwidth]{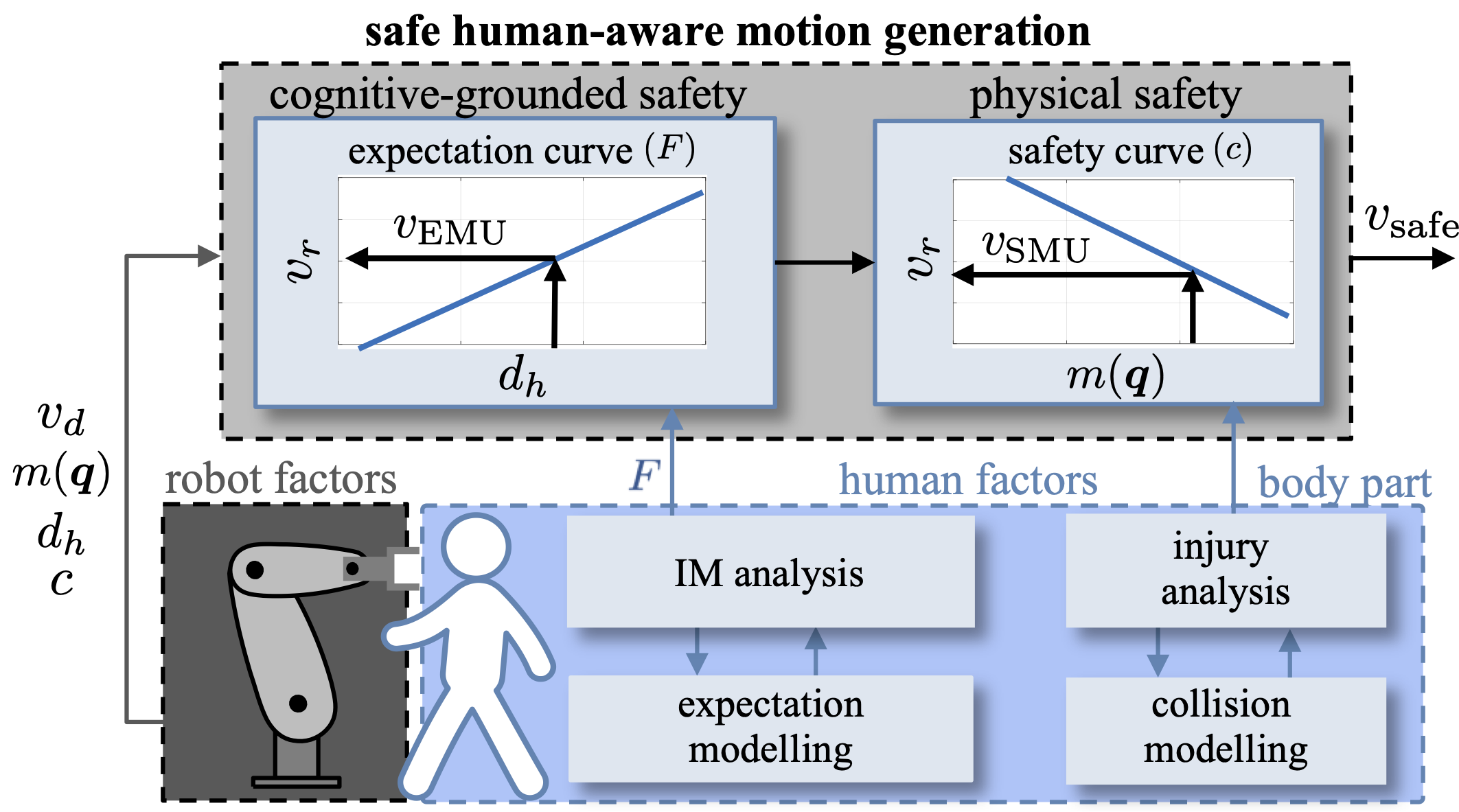}
	\caption{Safe human-aware motion generation scheme comprising the SMU and EMU. The framework's inputs are the desired robot velocity $v_d$, the instantaneous robot effective mass that depends on the robot pose $m(\vq)$, the current human-robot distance $d_h$, the endangered body part, the end-effector curvature $c$, and the human cognitive condition $F$. The output of the framework is a velocity that satisfies the physical and expectation-based safety constraints.}
	\label{fig:ablauf} 
\end{figure}

\subsection{EMU Implementation and Validation}
\begin{figure}
    \centering
    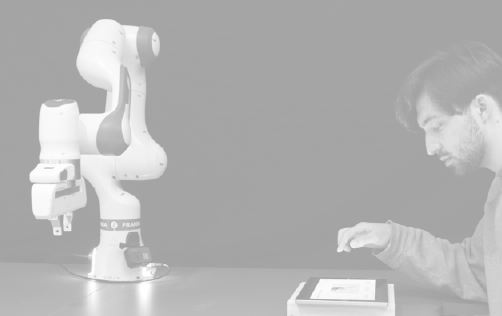
    \caption{EMU validation experiment: The desired robot velocity for maximum task performance $v_\mathrm{d}$ is shown in black; the blue line represents actual robot velocity $v_\mathrm{r}$ when applying the combination of EMU and SMU.}
    \label{fig:EMU-control}
    \vspace{-2mm}
\end{figure}

The EMU is cascaded with the Safe Motion Unit \cite{Haddadin_2012}. The framework is straightforward to implement on most commercial robot systems. On the Franka Emika Panda, we use the manufacturer's control interface and joint velocity controller. For the practical realization of the EMU, we select a linear expectation curve for the IMO threshold $q_r = 0.15$. Using this expectation curve, the instantaneous human-robot distance $d_\mathrm{h}$ is mapped to the safe velocity $v_\mathrm{EMU}$; see Fig. \ref{fig:s_curve}. This is done for robot-human distances of less than \SI{30}{cm}. The velocity limit provided by the EMU is then forwarded to the SMU, which checks whether the desired speed also satisfies the physical safety constraint. The SMU maps the current configuration-dependent robot reflected mass in the direction of motion $m(\vq)$ to a biomechanically safe velocity $v_\mathrm{SMU}$ via a curvature-related human injury threshold called safety curve; see Fig. \ref{fig:ablauf}. 
If the instantaneous robot-human distance is $d_\mathrm{h} \le d_\mathrm{max}$, the commanded robot velocity is the smallest of the three speeds $v_d$, $v_\mathrm{EMU}$, and $v_\mathrm{EMU}$. For distances $> d_\mathrm{max}$, the desired velocity is only limited by the SMU. Finally, the output of the framework is the safe robot velocity $v_\mathrm{safe}$ that prevents human injury and involuntary motion:
\begin{equation}
    v_\mathrm{safe} = \begin{cases}
    \min\{v_d, v_\mathrm{SMU}, v_\mathrm{EMU}\}, & \quad d_\mathrm{h} \le d_\mathrm{max} \\
    \min\{v_d, v_\mathrm{SMU}\}, & \quad d_\mathrm{h} > d_\mathrm{max} \\
\end{cases} 
\label{eq:}%
\end{equation}
To validate whether the EMU concept reduces IMO in practice, we repeat the experiment described in Sec. \ref{sec:exp_proc}, this time using EMU velocity shaping. The experimental setup and velocity profiles are depicted in Fig. \ref{fig:EMU-control}. The desired nominal velocity profile is shown in black, the actual robot velocity shaped by the combination of EMU and SMU in blue. Twenty participants are part of the validation experiment with an average age of $36 \pm 17.7$ years, including ten males ($50$~\%) and ten females ($50$~\%). We want the relative frequency of IMO observed in this experiment to be less or equal to the desired threshold $q_r = 0.15$.  


The inter-coder reliability for the validation experiment is $\kappa =$ 0.63, which can be considered as ``substantial'' according to \cite{Landis_1977}. Compared to the previous experiment, the lower $\kappa$ score for the inter-coder reliability may result from less strong S-S cues (which are difficult to identify) and the lower number of participants. 
Due to the lower inter-coder reliability, both coders' results are used to validate the EMU. Table \ref{tab:final_results} lists the relative frequency $f$ of S-S cues rated by the first (C1) and second coder (C2). 
%
%
%
\begin{table}[]
    \centering
    \caption{Rel. frequencies $f$ of S-S cues depending on robot distance and velocity, evaluated by coders C1 and C2 in the validation study}
    \begin{center}
    \begin{tabular}{cccc}
        \toprule
        $d_\mathrm{h}~$[m] & $v_\mathrm{r}~$[m/s] & $f$ by C1 & $f$ by C2  \\
        \midrule
        0.25 & 0.405 & 0.03 & 0.00 \\
        0.20 & 0.33 & 0.05 & 0.05 \\
        0.15 & 0.255 & 0.00 & 0.03 \\
        0.10 & 0.18 & 0.07 & 0.03 \\
        0.05 & 0.105 & 0.21 & 0.13 \\
        0.00 & 0.03 & 0.23 & 0.36 \\
        \bottomrule
    \end{tabular}
    \end{center}
    \label{tab:final_results}
    \vspace{-5mm}
\end{table}
From the evaluation of both coders, we conclude that in the case of an approach to a distance $d_h >$ \SI{10}{cm}, the relative frequency of S-S occurrence is very low ($0-5$~\%). For robot-human distances $d_h <$ \SI{10}{cm} the risk of IMO is $3-7$~\%. Below \SI{5}{cm}, $13-21$~\% IMO is observed. The difference of $8$~\% in the relative frequency between both coders indicates that the reaction was rated differently in three cases. In our preliminary validation, we observe that the desired $15$~\% threshold was satisfied for robot-human distances $>$ \SI{5}{cm}. However, for $d_h \ge$ \SI{0}{cm} we observe $24-36$~\% IMO. A comparison of the number of IMO in the validation experiment (with EMU) and the previous experiment (without EMU) in terms of a one-sided t-test under the assumption of unequal variances ($p=0.005$) yields that the EMU significantly reduces IMO. For the first group of participants, we could not prove a normal distribution according to the Kolmogorov-Smirnov test at the significance niveau $\alpha = 0.05$. However, the t-test was performed under the assumption that it is robust for non-normal distributed data.



To sum up, the EMU generates motions that result in the desired reduction in IMO in five of six robot approaches. For human-robot distances $\ge \SI{10}{cm}$, the robot velocity can even be increased while ensuring the IMO constraint. However, in the immediate vicinity of the human body, which was on the boundary of the human workspace ($\approx$ \SI{4}{cm} from the human fingertips), IMO was not reduced sufficiently.




\section{DISCUSSION}
\label{sec:disc}
Once a set of risk matrices for IMO is established, IMO thresholds need to be set to obtain expectation curves. The thresholds may either be selected globally, e.g., according to the probability of the human intruding the robot's workspace or vice versa \cite{Kim_2020}, or dynamically depending on the human condition. For example, one may select a rather low threshold $q_r$ when the user is sleepy and a higher threshold when the human observes the robot task closely. To deploy the EMU approach in real application scenarios, we need to
\begin{enumerate}
	\item identify and monitor the human condition using a human profiler (eye-tracking can be used to determine the human's level of awareness, for example),
	\item select the risk matrix based on a scenario and human condition, and 
	\item define the desired IMO probability threshold and the respective expectation curve.
\end{enumerate}

In this letter, we used social signal coding to evaluate human IMO. Future work should also include objective criteria such as EEG measurement on the frontal belly of the epicranius muscle or skin conductance measurements. As mentioned in Sec. IV, heart rate and heart rate variability did not correlate with IMO in our experiments, unfortunately.

Human IMO depends on several human, robot, and environmental factors. This work systematically investigates the relationship between the robot velocity and human-robot distance and human involuntary motion occurrence in a common HRI scenario. The influence of further robot and human factors should be investigated in future work. Concerning the age group 60+ in our experiments, the two coders classified only one out of 72 approaches as an S-S cue. This suggests that high age (which may lead to decreased reactivity, e.g.) is a human factor that influences IMO. We will dive deeper into analyzing particular human factors in our future research. Furthermore, different robot types, HRI scenarios, and motion parameters (e.g., acceleration, jerk) are worth investigating. However, our systematic and generalizable approach from the analysis of IMO to safe, real-time capable robot motion generation allows integrating future results in a straightforward manner; the principle structure of the EMU remains the same.

\section{CONCLUSION}
\label{sec:conc}
In this letter, we introduced, elaborated, and validated the Expectable Motion Unit (EMU) concept, which aims at avoiding possibly hazardous human involuntary motions (IM) in human-robot interaction. We conducted experiments with 29 volunteers to systematically analyse the relative frequency of IM occurrence (IMO) depending on the robot speed and human-robot distance in a common HRI setting. The experimental results are processed towards a risk matrix, from which expectation curves were deduced for a particular application. An expectation curve that limits the probability of IMO to \SI{15}{\%} in the considered use case was embedded into the EMU motion generator, which limits the robot velocity such that the IM threshold is not exceeded. Furthermore, the EMU was combined with the well-established Safe Motion Unit \cite{Haddadin_2012} that ensures physical safety during contact. In a validation experiment, the EMU successfully prevented IM in five out of six cases. Overall, by fulfilling the human expectation of the robot behavior and taking the biomechanical safety limits into account at the same time, our concept improves the safety, performance, and trust of robot users in HRI.

\section*{ACKNOWLEDGMENT}
The authors would like to thank Simone Stahl, Jan Schreiber, Melanie Porzenheim, Leon Oskui, and UAS Fulda for supporting the experiments. 
Please note that S. Haddadin has a potential conflict of interest as shareholder of Franka Emika GmbH.

		    





\bibliographystyle{IEEEtran} 
\bibliography{bib}


\end{document}